\documentclass[runningheads]{llncs}

\usepackage{eccv}

\usepackage{eccvabbrv}
\usepackage{wrapfig}
\usepackage{multirow}
\usepackage{graphicx}
\usepackage{booktabs}
\usepackage{caption}
\usepackage{comment}

\usepackage{colortbl} 
\definecolor{grayc}{gray}{0.9} 
\usepackage[table]{xcolor}
\definecolor{lightred}{RGB}{255, 230, 230} 

\usepackage[accsupp]{axessibility}  
\usepackage{enumitem}

\usepackage{hyperref}

\usepackage{orcidlink}

\begin{document}

\title{RefReward-SR: LR-Conditioned Reward Modeling for Preference-Aligned Super-Resolution} 

\titlerunning{RefReward-SR}

\author{Yushuai Song\inst{1,2} \and
Weize Quan\inst{1,2} \and
Weining Wang\inst{1,2} \and
Jiahui Sun\inst{1,2} \and
Jing Liu\inst{1,2} \and
Meng Li\inst{3} \and
Pengbin Yu\inst{3} \and
Zhentao Chen\inst{3} \and
Wei Shen\inst{3} \and
Lunxi Yuan\inst{3} \and
Dong-ming Yan\inst{1,2}}

\authorrunning{F.~Author et al.}

\institute{Institute of Automation, Chinese Academy of Sciences \and
School of Artificial Intelligence, University of Chinese Academy of Sciences \and
OPPO AI Center, OPPO Inc.\\
\email{\{songyushuai2024\}@ia.ac.cn}}

\maketitle

\begin{abstract}
    Recent advances in generative super-resolution (SR) have greatly improved visual realism, yet existing evaluation and optimization frameworks remain misaligned with human perception. Full-Reference and No-Reference metrics often fail to reflect perceptual preference, either penalizing semantically plausible details due to pixel misalignment or favoring visually sharp but inconsistent artifacts. Moreover, most SR methods rely on ground-truth (GT)–dependent distribution matching, which does not necessarily correspond to human judgments. In this work, we propose RefReward-SR, a low-resolution (LR) reference-aware reward model for preference-aligned SR. Instead of relying on GT supervision or NR evaluation, RefReward-SR assesses high-resolution (HR) reconstructions conditioned on their LR inputs, treating the LR image as a semantic anchor. Leveraging the visual–linguistic priors of a Multimodal Large Language Models (MLLM), it evaluates semantic consistency and plausibility in a reasoning-aware manner. To support this paradigm, we construct RefSR-18K, the first large-scale LR-conditioned preference dataset for SR, providing pairwise rankings based on LR–HR consistency and HR naturalness. We fine-tune the MLLM with Group Relative Policy Optimization (GRPO) using LR-conditioned ranking rewards, and further integrate GRPO into SR model training with RefReward-SR as the core reward signal for preference-aligned generation. Extensive experiments show that our framework achieves substantially better alignment with human judgments, producing reconstructions that preserve semantic consistency while enhancing perceptual plausibility and visual naturalness. Code, models, and datasets will be released upon paper acceptance. 
\end{abstract}

\section{Introduction}
\label{sec:intro}

Image Super-Resolution (SR) \cite{dong2014learning,lim2017enhanced,ledig2017photo,zhang2018image,wang2018esrgan,liang2021swinir,wang2021real} aims to reconstruct high-resolution (HR) images from low-resolution (LR) observations. As an inherently ill-posed inverse problem, SR has evolved from bicubic degradation settings to real-world scenarios with complex and unknown degradations\cite{moser2024diffusion}. The ultimate goal of real-world SR is to produce perceptually pleasing results that align with human visual aesthetics while preserving content fidelity. Early representative methods, such as BSRGAN \cite{zhang2021designing} and Real-ESRGAN \cite{wang2021real}, formulated this task as a supervised distribution-matching problem, leveraging adversarial losses to synthesize plausible high-frequency details. With diffusion priors\cite{SD2022,flux2024}, recent methods\cite{wang2024exploiting,lin2024diffbir,yang2024pixel,ai2024dreamclear,wu2024seesr,tai2026addsr} surpass GAN-based models, generating fine details even under severe degradation.

However, as generative models become increasingly powerful, existing evaluation and optimization frameworks for SR are encountering significant limitations. We summarize these challenges into two critical aspects:

\textbf{1) Inadequacy of Existing Metrics for Semantic Alignment.}
Traditional Full-Reference (FR) metrics \cite{wang2004image,zhang2018unreasonable} provide indispensable benchmarks for measuring pixel-level or structural fidelity. However, their reliance on strict spatial correspondence can be overly conservative for generative SR, where they may penalize ``reasonable hallucinations''—details that are semantically sound and visually enriching but spatially misaligned with the ground-truth (GT) \cite{blau2018perception}.
On the other hand, No-Reference (NR) metrics \cite{yang2022maniqa,ke2021musiq,chen2024topiq,wang2023exploring,wu2023q,you2025teaching,wu2025visualquality} excel at assessing low-level visual sharpness and textural naturalness without requiring a reference. However, due to the lack of reference, they inherently struggle to preserve content fidelity. This often allows models to ``game the metric'' by introducing sharp but unfaithful artifacts, or even altering the subject identity, while still receiving high scores.
More importantly, both FR and NR metrics evaluate high-level semantic plausibility only passively through low-level cues, ignoring the top-down cognitive mechanism where human perception naturally prioritizes semantic consistency before scrutinizing low-level details. Thus, they fail to explicitly penalize obvious semantic violations or structural distortions that contradict world knowledge. Consequently, existing metrics struggle to fully align with true human preferences.


\textbf{2) The Insufficiency of Exclusively GT-Driven Optimization.}
Most SR methods \cite{wang2024exploiting,lin2024diffbir,yang2024pixel,wu2024seesr} rely on supervised losses for fidelity, while advanced approaches \cite{wu2025omgsr,lin2025harnessing} incorporate adversarial losses. However, while GT supervision provides necessary structural bounds, relying exclusively on GT distribution matching restricts generative potential. Strictly fitting empirical GT is insufficient to capture higher-level human semantic preferences. Furthermore, real-world GT sometimes contains inherent degradations, causing overly GT-dependent models to reproduce imperfections rather than enhance quality. Unlike the Text-to-Image field, which actively aligns with human preferences via RLHF \cite{xu2023imagereward,black2023training,wallace2024diffusion}, using RLHF to complement GT constraints in SR remains limited.

\begin{figure}[t]
\centering
\includegraphics[width=\textwidth]{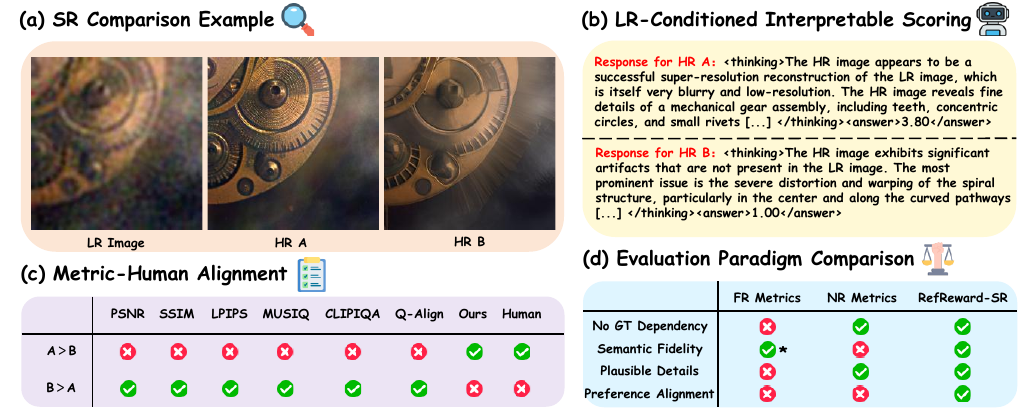} 
\caption{Overview of our proposed evaluation paradigm. (a) HR A preserves structures, while HR B shows severe distortions. (b) RefReward-SR uses an MLLM for interpretable, LR-conditioned semantic scoring. (c) Existing metrics wrongly favor the distorted HR B, whereas ours aligns with human perception. (d) RefReward-SR uniquely ensures semantic fidelity and preference alignment without GT dependency. The asterisk(*) indicates that FR metrics evaluate high-level semantics only passively via low-level correspondence.}
\label{fig:teaser}
\vspace{-10pt}
\end{figure}

In this work, we propose \textbf{RefReward-SR}, an LR reference-aware semantic scoring model designed to complement existing evaluation paradigms. Addressing the ill-posed nature of super-resolution, RefReward-SR mimics human assessment by treating the LR image as a semantic anchor, leveraging the visual-linguistic priors of a Multimodal Large Language Model (MLLM) to jointly assess semantic consistency and plausibility from a high-level cognitive perspective. To this end, we construct \textbf{RefSR-18K}, the first large-scale LR-conditioned preference dataset focused on semantic-level annotations, establishing a critical foundation to shift SR evaluation and optimization from pixel-matching to human preference alignment. We fine-tune the MLLM using Group Relative Policy Optimization (GRPO)\cite{shao2024deepseekmath,deepseek2025r1} to align its judgments with human preferences, thereby capturing the underlying logical reasoning intrinsic to human perception. This enables RefReward-SR to serve as a key semantic complement to low-level metrics, establishing a more comprehensive evaluation and optimization framework for generative SR.

Building upon the learned reward models, we introduce GRPO into the super-resolution framework, employing a joint reward function with RefReward-SR as the core semantic signal to guide preference-aligned generation. Experimental results demonstrate that incorporating RefReward-SR alongside perceptual and quality constraints significantly enhances alignment with human judgments, leading to reconstructions with superior semantic consistency and plausibility.

Our contributions can be summarized as follows:

\begin{itemize}

    \item We construct the first large-scale LR-conditioned preference dataset for SR. By treating the LR image as a semantic anchor, RefSR-18K provides 18,000 pairwise ranking annotations across diverse images, prioritizing high-level semantic consistency and plausibility.

    \item We propose RefReward-SR, an LR reference-aware semantic scoring model explicitly aligned with human perception. To address fine-grained discrimination, we further introduce a \emph{global-local crop scoring} strategy to enhance sensitivity to localized semantic violations and artifacts.

    \item We introduce GRPO into the SR task with RefReward-SR as the reward signal, enabling preference-aligned optimization that improves semantic consistency and visual plausibility.

\end{itemize}

\section{Related Work}

\subsection{Evaluation Metrics for Super-Resolution}
Traditional FR metrics, including PSNR, SSIM \cite{wang2004image}, and LPIPS \cite{zhang2018unreasonable}, quantify reconstruction quality by measuring deviations from the ground truth (GT). While indispensable for ensuring basic structural fidelity, operating within a strict distortion-minimization framework \cite{blau2018perception} causes them to systematically penalize spatially misaligned yet perceptually plausible details. Moreover, the frequent unavailability of high-quality GT in real-world scenarios further limits their practicality \cite{chen2025toward}.
To remove GT dependency, NR metrics such as MUSIQ\cite{ke2021musiq}, MANIQA\cite{yang2022maniqa}, and CLIPIQA\cite{wang2023exploring} assess image quality without supervision. Recent IQA approaches built on Multimodal Large Language Models (MLLMs)\cite{wu2023q,you2025teaching,wu2025visualquality} introduce partial interpretability and region awareness. Although effective at evaluating low-level visual aesthetics, their lack of a semantic anchor inherently causes them to struggle to preserve content consistency, often allowing models to inflate scores via excessive sharpening or unfaithful hallucinations.
Furthermore, neither FR nor NR metrics can systematically evaluate semantic plausibility, which is the primary focus of human assessment for SR results.
In contrast, we formulate SR evaluation as an LR-conditioned preference modeling problem. By jointly reasoning over the LR input and its HR reconstruction, RefReward-SR leverages the world knowledge of a Multimodal Large Language Model to provide semantically grounded, LR-anchored, and preference-aligned evaluation signals, effectively complementing the limitations of traditional FR/NR paradigms in semantic-level human preference assessment.

\subsection{Generative Image Super-Resolution}
Early SR methods optimized pixel-wise reconstruction losses under predefined degradations, achieving high PSNR but producing over-smoothed results \cite{dong2014learning,lim2017enhanced,kim2016accurate,lai2017deep,tong2017image,zhang2018rcan}. To improve visual quality, adversarial learning was introduced. Models such as ESRGAN\cite{zhang2021designing} and Real-ESRGAN\cite{wang2021real} framed SR as distribution matching, generating high-frequency details via GAN objectives. Later, diffusion-based approaches\cite{lin2024diffbir,wu2024seesr,lin2025harnessing,wu2024one} leveraging pretrained models like Stable Diffusion\cite{SD2022} and FLUX\cite{flux2024} further advanced generative SR, demonstrating stronger semantic priors and improved perceptual quality under complex degradations. Despite these advances, existing optimization paradigms remain focused on fitting the GT data distribution, rather than directly optimizing objectives aligned with human preferences. Moreover, they are fundamentally constrained by the quality of GT images. To address this limitation, inspired by reinforcement learning in text-to-image generation\cite{xu2023imagereward,black2023training,wallace2024diffusion,xue2025dancegrpo,liu2025flow,guo2025imagedoctor}, several works\cite{cai2025dspo,wu2025dp,qiao2025realsr,xu2025irpo} have begun exploring reinforcement learning for SR. However, these methods often rely on existing evaluation metrics\cite{cai2025dspo,wu2025dp} with inherent limitations, or on MLLMs\cite{qiao2025realsr,xu2025irpo} that are not well-suited for low-level vision tasks, preventing them from fully unlocking the potential of reinforcement learning in SR.

\section{Method}
\subsection{Construction of the RefSR-18K Dataset}
\label{sec:dataset}

\noindent\textbf{Motivation.} In real-world generative SR, models frequently exhibit similar failure modes, such as structural distortions, unreasonable textures, and localized content hallucinations that alter the semantics of the LR input. Traditional Image Quality Assessment (IQA)\cite{sheikh2006statistical,lin2019kadid,fang2020perceptual} and aesthetic datasets\cite{xu2023imagereward,kirstain2023pick,wu2023human} fail to capture these generative anomalies, as they primarily focus on low-level distortions (e.g., noise) or global artistic appeal at the single-image level. Consequently, they lack corresponding LR references to evaluate semantic consistency and fail to provide explicit rankings for the semantic plausibility of the SR results. To address this gap, we construct RefSR-18K, a preference dataset specifically tailored for generative SR. Conditioned on the LR input, this dataset comprehensively evaluates the reconstructed HR images from two perspectives: semantic consistency and semantic plausibility. This provides crucial data support for training LR-conditioned, semantic-aware evaluation models.

\noindent\textbf{Data Collection and Candidate Generation.}
To ensure diversity and broad scene coverage, we curated 5,130 high-quality images from the LSDIR dataset\cite{li2023lsdir}, encompassing a wide range of visual categories. Following the degradation pipeline proposed in SeeSR\cite{wu2024seesr}, we applied random cropping and complex degradations to synthesize the corresponding LR inputs. To ensure comprehensiveness and broad representativeness, the dataset must encompass diverse reconstruction styles and typical generative distortions produced by various models. To achieve this, we employed 8 diverse, state-of-the-art SR models to generate HR candidates: DiffBIR\cite{lin2024diffbir}, SeeSR\cite{wu2024seesr}, OESDiff\cite{wu2024one}, S3Diff\cite{2024s3diff}, LucidFlux\cite{fei2025lucidflux}, DiT4SR\cite{duan2025dit4sr}, OMGSR-S\cite{wu2025omgsr}, and HYPIR\cite{lin2025harnessing}. This model pool deliberately covers both single-step and multi-step generation paradigms, as well as distinct generative priors based on Stable Diffusion\cite{SD2022} and FLUX\cite{flux2024} architectures.

\noindent\textbf{Annotation Protocol and Quality Control.}
Each annotation group consists of one LR reference image and four HR candidates randomly sampled from a pool of eight model outputs and the GT. Annotators are instructed to rank the four candidates based on two primary criteria: 
(1) \textbf{LR--HR semantic consistency}, i.e., whether the generated details faithfully align with the structures and semantic cues present in the LR input; and 
(2) \textbf{semantic plausibility}, i.e., whether reconstructed objects are semantically natural and free from distortion-induced anomalies.
When quality differences are indistinguishable, tied rankings are allowed. Detailed annotation guidelines and visual examples are provided in the \textit{Suppl. Mater.}.
Given the subjective and fine-grained nature of human preference annotation in image restoration, we adopt a strict quality control pipeline. Annotators are required to complete training and qualification tests prior to participation, and only those demonstrating high agreement with expert judgments are retained (approximately 50\% are filtered out). Each group is independently annotated by at least three qualified annotators, and the final ranking is obtained via average rank aggregation. We further conduct expert review and post-hoc filtering to remove unreliable samples, including cases where candidate qualities are nearly indistinguishable, images lack sufficient semantic content for evaluation, or substantial disagreement exists among annotators. After this rigorous filtering process, 4,699 high-quality annotation groups are retained, resulting in 18,796 annotated HR reconstructions paired with LR references, forming the RefSR-18K dataset.

\subsection{RefReward-SR: A Reference-Aware MLLM Evaluator}
\label{sec:refreward}
\begin{figure*}[t]
\centering
\includegraphics[width=1\textwidth]{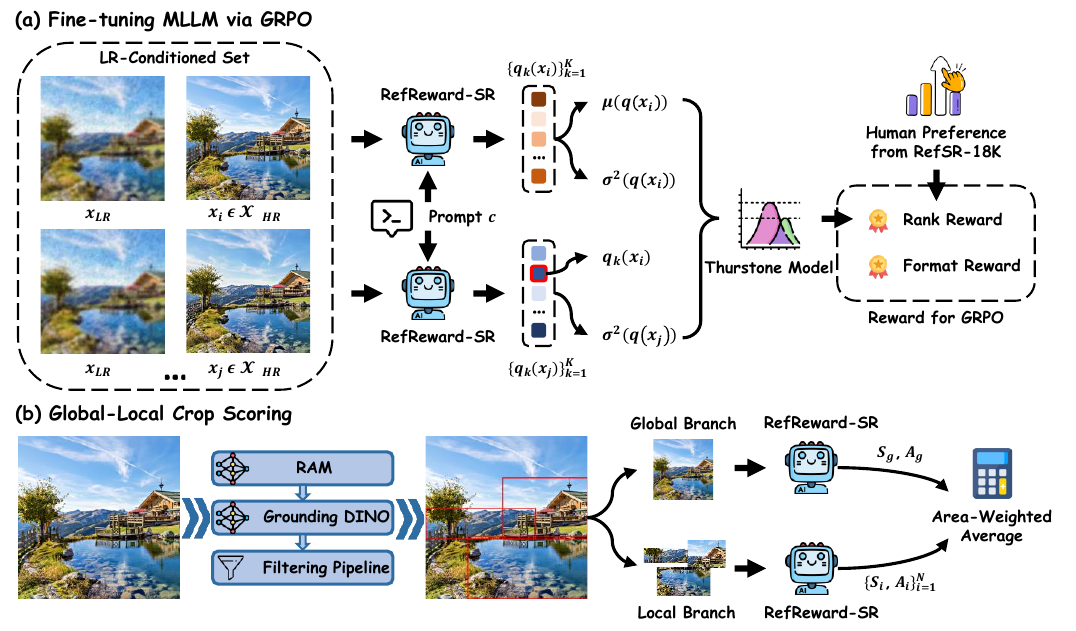} 
\caption{\textbf{The proposed RefReward-SR framework.} (a) \textbf{MLLM Fine-tuning}: Fine-tuning via GRPO with format and LR-conditioned rank rewards from RefSR-18K for human preference alignment. (b) \textbf{Global-Local Crop Scoring}: Extracting representative local crops via RAM and Grounding DINO, then fusing multi-scale MLLM scores via area-weighted averaging for comprehensive evaluation.}
\label{fig:method}
\end{figure*}
To build a human-aligned, reference-aware semantic evaluator, we fine-tune Qwen3-VL 8B~\cite{Qwen3-VL} via GRPO~\cite{shao2024deepseekmath,deepseek2025r1}. Given a low-resolution image $x_{LR}$, an evaluation prompt $c$ specifying the scoring criteria and reasoning instructions, and a high-resolution candidate $x_i$ from a candidate set $\mathcal{X}_{HR}$ (where $|\mathcal{X}_{HR}|=4$) in the RefSR-18K dataset, the MLLM generates a detailed reasoning process followed by a structured score. The total reward for an output $O_i$ comprises two core components:

\noindent\textbf{Format Reward.}
\label{sec:format_reward}
To ensure the model functions as \emph{an interpretable evaluator} rather than a ``black-box'' scorer, we enforce a structured output format. By inducing reasoning through reinforcement learning, we encourage the model to automatically explore reasonable reasoning paths guided by the evaluation prompt $c$. The model must first generate a reasoning chain within \texttt{$<$thinking$>$}\allowbreak\dots\allowbreak\texttt{$<$/thinking$>$} tags, explicitly articulating the issues present in the HR image and their specific locations conditioned on the LR reference. The final numerical rating is then enclosed in \texttt{$<$answer$>$\dots$<$/answer$>$} tags. If any of these formats are incorrect, the format score is $0$. The reward is defined as:
\begin{equation}
    R_{\text{format}}(O_i) = 
    \begin{cases} 
      1.0, & \text{if } O_i \text{ satisfies all format requirements} \\ 
      0, & \text{otherwise} 
    \end{cases}
    \label{eq:format_reward}
\end{equation}

\noindent\textbf{LR-Conditioned Rank Reward.}
\label{sec:rank_reward}
The core of RefReward-SR is to learn the relative quality ordering of HR candidates conditioned on the same LR anchor $x_{LR}$ and the evaluation prompt $c$. For a set of $G$ HR candidates $\mathcal{X}_{HR} = \{x_1, \dots, x_G\}$ corresponding to a specific LR input $x_{LR}$, we apply GRPO to generate $K$ quality score predictions for each candidate $x_i$ under the joint condition $(x_{LR}, c)$, denoted as $\{q_k(x_i \mid x_{LR}, c)\}_{k=1}^{K}$. Here, $q_k(x_i \mid x_{LR}, c)$ denotes the scalar quality score produced by the $k$-th stochastic rollout of the policy $\pi_\theta(\cdot \mid x_i, x_{LR}, c)$. For notational simplicity, the conditioning on $(x_{LR}, c)$ is implicit in the following formulations.

Based on the Thurstone model \cite{thurstone2017law,wu2025visualquality}, we compute the comparative probability of the $k$-th prediction of $x_i$ against another candidate $x_j$ within the same set by leveraging their sample means and variances:
\begin{equation}
    p_k(x_i, x_j) = \Phi \left( \frac{q_k(x_i) - \mu(q(x_j))}{\sqrt{\sigma^2(q(x_i)) + \sigma^2(q(x_j)) + \gamma}} \right)
    \label{eq:prob_compare},
\end{equation}
where $\mu(\cdot)$ and $\sigma^2(\cdot)$ are the sample mean and variance of the $K$ predictions, $\Phi$ is the standard normal cumulative distribution, and $\gamma$ is a small stabilizing constant. This explicitly accommodates predictive uncertainty for different images.

The true preference $p(x_i, x_j) \in \{1, 0.5, 0\}$ is derived from human annotations in RefSR-18K (representing win, tie, or loss). We define the reward $R_k(x_i)$ using a continuous fidelity measure averaged across the remaining $G-1$ candidates:
\begin{equation}
    R_k(x_i) = \frac{1}{G-1} \sum_{j \neq i} \left( \sqrt{p(x_i, x_j)p_k(x_i, x_j)} + \sqrt{(1 - p(x_i, x_j))(1 - p_k(x_i, x_j))} \right)
    \label{eq:rank_reward}.
\end{equation}
This continuous reward provides precise guidance by capturing subtle variations in quality ranking, penalizing outlier predictions.

\noindent\textbf{Global-Local Crop Scoring.}
\label{sec:crop_scoring}
In practice, different SR results often exhibit consistent overall structures, with quality differences primarily manifesting in local details. Consequently, relying solely on global scoring can easily cause these critical nuances to be overlooked. During manual annotation, human evaluators typically observe the entire image first before comparing the primary objects individually. To mimic this behavior, we propose Global-Local Crop Scoring, which performs a holistic assessment followed by a fine-grained local comparison.

The strategy operates in two stages: (1) \textbf{Region Proposals}: We first leverage the RAM\cite{zhang2023recognize} to extract semantic tags, which then guide Grounding DINO\cite{liu2023grounding} to generate candidate bounding boxes. To mimic human visual behavior that prioritizes texture-rich and salient objects, while balancing evaluation efficiency, we design a bounding box filtering pipeline to select representative regions with low overlap (details in \textit{Suppl. Mater.}). (2) \textbf{Joint Evaluation}: The MLLM evaluates the global image and the selected crops separately. To integrate these multi-scale assessments, the final score $S_{\text{final}}$ is computed as an area-weighted average:
\begin{equation}
    S_{\text{final}} = \frac{A_g S_g + \sum_{i=1}^{N} A_i S_i}{A_g + \sum_{i=1}^{N} A_i}
    \label{eq:area_weight},
\end{equation}
where $S_g, A_g$ denote the score and area of the global image, and $S_i, A_i$ represent those of the $i$-th local crop. This mechanism accounts for both global information and local details, leading to a more accurate assessment score.

\subsection{Preference-Aligned Super-Resolution via GRPO}
\label{sec:sr_grpo}

Having established RefReward-SR as a reliable human-aligned evaluator, 
we further integrate it into the SR generation stage 
to directly optimize reconstruction quality toward human preference alignment. 
In this work, we adopt the C-FLUX model from DP$^2$O-SR~\cite{wu2025dp} as our base generative SR model. 
Specifically, we formulate the fine-tuning of this C-FLUX model as a reinforcement learning problem, 
which is optimized using GRPO\cite{liu2025flow,xue2025dancegrpo}.

Let the pre-trained SR diffusion model be parameterized as a conditional policy 
$\pi_\theta$, which generates a high-resolution reconstruction 
$y \sim \pi_\theta(\cdot \mid x_{LR})$ conditioned on a low-resolution input $x_{LR}$.
For each input $x_{LR}$, the policy produces a group of stochastic reconstructions 
$\mathcal{Y} = \{y_1, y_2, \dots, y_G\}$, and optimization is performed based on 
relative quality comparisons within each group. 
Since the learning signal is entirely driven by reward feedback, 
the design of the reward function becomes the key factor for achieving preference alignment.
To comprehensively enhance SR reconstruction quality, we construct a composite reward consisting of three complementary components.

\noindent\textbf{RefReward-SR ($R_{\text{Ref}}$).}
The primary reward signal is provided by RefReward-SR, our LR reference-aware evaluator described in Sec.~\ref{sec:refreward}.
It jointly assesses semantic consistency between the generated image $y_i$ and the LR observation $x_{LR}$, as well as the semantic plausibility of the reconstructed content. 

\noindent\textbf{LPIPS Perceptual Reward ($R_{\text{LPIPS}}$).}
To prevent excessive structural deviation and maintain perceptual fidelity with respect to the ground-truth image $y_{\text{GT}}$, 
we incorporate the LPIPS~\cite{zhang2018unreasonable} metric as a perceptual constraint. 
Since lower LPIPS values indicate higher perceptual similarity, it is incorporated as a negative penalty term.

\noindent\textbf{DEQA Quality Reward ($R_{\text{DEQA}}$).}
We further incorporate DEQA~\cite{you2025teaching} as an image quality assessment reward to enhance overall visual sharpness and strengthen the penalization of low-level visual degradations (e.g., blur and noise).

The overall reward for each candidate reconstruction $y_i$ is defined as

\begin{equation}
R_{\text{total}}(y_i)
=
\lambda_1 R_{\text{Ref}}(y_i, x_{LR})
-
\lambda_2 R_{\text{LPIPS}}(y_i, y_{\text{GT}})
+
\lambda_3 R_{\text{DEQA}}(y_i),
\label{eq:total_reward_grpo}
\end{equation}
where $\lambda_1$, $\lambda_2$, and $\lambda_3$ are hyperparameters balancing semantic
human preference alignment, perceptual fidelity, and artifact suppression, respectively.

\section{Experiments}
\label{sec:Experiments}
\subsection{Effectiveness of RefReward-SR}
\label{sec:refreward-result}
To validate whether RefReward-SR can assess SR results at the semantic level in alignment with human judgment, and to investigate what extent existing evaluation metrics capture human preferences, we conduct a comprehensive evaluation mimicking human selection behavior and compare our model against traditional NR/FR metrics and state-of-the-art MLLMs.

\begin{table}[ht]
\centering
\vspace{-0.3cm}
\caption{Quantitative results (in \%) on human preference alignment. We report agreement with human annotators, as well as Recall@1 and Filter@1 on both in-domain and out-of-domain test sets. ``Annotators'' denotes the average agreement among human annotators. All $\pm$ values indicate the maximum deviation from the mean.}
\label{tab:main_results}
\resizebox{0.9\textwidth}{!}{
\begin{tabular}{lcccccc}
\toprule
Method & \multicolumn{3}{c}{In-Domain} & \multicolumn{3}{c}{Out-of-Domain} \\
\cmidrule(lr){2-4} \cmidrule(lr){5-7}
& Agreement & Recall@1 & Filter@1 & Agreement & Recall@1 & Filter@1 \\
\midrule
\textit{Human Agreement} \\
Annotators & $84.7 \pm 1.4$ & - & - & $81.8 \pm 1.3$ & - & - \\
\midrule
\textit{FR Metrics} \\
PSNR & $41.8 \pm 1.6$ & 45.0 & 9.5 & $35.5 \pm 2.3$ & 25.5 & 8.5 \\
SSIM~\cite{wang2004image} & $51.8 \pm 2.0$ & 51.0 & 25.5 & $43.3 \pm 1.8$ & 26.0 & 19.0 \\
LPIPS~\cite{zhang2018unreasonable} & $62.5 \pm 0.9$ & 49.5 & 46.0 & $58.8 \pm 2.0$ & 44.5 & 34.0 \\
\midrule
\textit{NR Metrics} \\
MUSIQ~\cite{ke2021musiq} & $60.9 \pm 1.4$ & 40.5 & 42.0 & $57.6 \pm 2.7$ & 38.0 & 34.5 \\
CLIPIQA~\cite{wang2023exploring} & $54.0 \pm 1.7$ & 31.0 & 30.5 & $56.9 \pm 2.5$ & 32.0 & 37.5 \\
NIMA~\cite{talebi2018nima} & $54.5 \pm 2.0$ & 31.5 & 24.0 & $59.5 \pm 2.3$ & 41.0 & 40.0 \\
TOPIQ-IAA~\cite{chen2024topiq} & $54.1 \pm 1.6$ & 36.0 & 27.0 & $56.4 \pm 1.7$ & 36.0 & 33.5 \\
Q-Align~\cite{wu2023q} & $54.9 \pm 1.7$ & 35.0 & 29.5 & $54.3 \pm 1.0$ & 40.5 & 32.0 \\
VQ-R1~\cite{wu2025visualquality} & $47.8 \pm 1.3$ & 49.0 & 48.0 & $44.1 \pm 2.8$ & 50.5 & 39.0 \\
\midrule
\textit{MLLMs} \\
GPT-5.2~\cite{achiam2023gpt} & $52.5 \pm 1.0$ & 30.7 & 46.2 & $49.7 \pm 2.2$ & 38.1 & 29.9 \\
Gemini 3 Pro~\cite{team2023gemini} & $58.8 \pm 1.5$ & 54.0 & 52.0 & $47.9 \pm 2.1$ & 42.6 & 36.6 \\
Qwen3-VL 8B~\cite{Qwen3-VL} & $31.8 \pm 0.4$ & 67.5 & 58.0 & $26.1 \pm 0.9$ & 59.0 & 57.0 \\
Qwen3-VL 32B~\cite{Qwen3-VL} & $46.4 \pm 1.3$ & 39.5 & 43.0 & $40.7 \pm 0.4$ & 38.5 & 36.0 \\
\midrule
\textbf{RefReward-SR} & \textbf{85.0 ± 2.4} & \textbf{84.5} & \textbf{78.0} & \textbf{80.2 ± 2.2} & \textbf{77.0} & \textbf{73.0} \\
\bottomrule
\end{tabular}
}
\vspace{-0.3cm}
\end{table}

\noindent\textbf{Experimental Setup.}
To evaluate performance and generalization, we establish two test sets. Both sets share the same 200 LR reference images but differ in the source of their HR candidates:
(1) \textbf{In-domain Test Set}: We randomly sample 200 groups directly from the original RefSR-18K dataset. These groups contain HR candidates generated by the 8 models seen during training, along with GT.
(2) \textbf{Out-of-domain Test Set}: Utilizing the identical 200 LR inputs from the In-domain set, we construct new comparison groups to assess generalization. In each group, 2 candidates are generated by entirely unseen models (TSD-SR~\cite{dong2025tsd} and PiSA-SR~\cite{sun2025pixel}), while the remaining 2 are sampled from the GT and the SR models used in training.
The out-of-domain set is annotated following the exact same pipeline as RefSR-18K to ensure consistency.

We compare RefReward-SR with three categories of evaluators:
(1) FR metrics, including PSNR, SSIM~\cite{wang2004image}, and LPIPS~\cite{zhang2018unreasonable};
(2) NR metrics, including MUSIQ~\cite{ke2021musiq}, CLIPIQA~\cite{wang2023exploring}, Q-Align~\cite{wu2023q}, NIMA~\cite{talebi2018nima}, VQ-R1~\cite{wu2025visualquality}, and TOPIQ-IAA~\cite{chen2024topiq}; and
(3) general-purpose MLLMs, including GPT-5.2~\cite{achiam2023gpt}, Gemini 3 Pro~\cite{team2023gemini}, and the non-fine-tuned Qwen3-VL 8B~\cite{Qwen3-VL}.
To ensure a fair comparison, all MLLMs are evaluated using the same prompt design as RefReward-SR.

Following the evaluation protocol established by ImageReward \cite{xu2023imagereward}, we assess the models based on agreement and their ability to successfully recall the best images or filter out the worst images. Specifically, \textit{Agreement} assesses the likelihood of the model sharing consistent preferences with human annotators. \textit{Recall@1} evaluates the model's ability to select the highly preferred images, measuring the frequency with which the human-annotated best image is ranked first by the model. Conversely, \textit{Filter@1} evaluates the model's ability to identify and penalize the worst generations by measuring how often the human-annotated worst image is ranked last. More details are provided in the \textit{Suppl. Mater.}.

\noindent\textbf{Results and Analysis.}
Table~\ref{tab:main_results} reports the quantitative results on semantic-level human preference alignment. RefReward-SR achieves performance comparable to human annotators on both in-domain and out-of-domain test sets, obtaining the best Recall@1 and Filter@1 while consistently outperforming all baselines. On the out-of-domain set, the performance of both human annotators and most metrics decreases, likely because the two unseen SR methods produce outputs with quality highly similar to other candidates, making preference discrimination more challenging. Although existing NR/FR metrics mainly focus on low-level features, they can indirectly reflect semantic-level human preferences, allowing most to maintain a positive agreement (i.e., >50\%). Notably, except for PSNR, all methods with agreement below 50\% are MLLM-based. When SR candidates share similar structures, zero-shot MLLMs tend to be insensitive to subtle image details and often assign identical scores. Under our calculation protocol (detailed in the \textit{Suppl. Mater.}), this behavior drives the agreement below 50\%, while paradoxically causing an upward trend in Recall@1 and Filter@1 (as typically seen in Qwen3-VL 8B). This phenomenon highlights the limitation of zero-shot MLLMs for fine-grained SR evaluation and further motivates the construction of the RefSR-18K dataset for preference-aligned fine-tuning.

\subsection{Performance of Preference-Aligned Super-Resolution}

\noindent\textbf{Training and Test Data.}
We construct our paired training data from the LSDIR dataset~\cite{li2023lsdir} using the Real-ESRGAN~\cite{wang2021real} degradation, setting the training resolution to $512\times512$ to align with our base model, C-FLUX~\cite{wu2025dp}. For evaluation, we follow the benchmark protocol provided by StableSR~\cite{wang2024exploiting}, including both synthetic and real-world test sets. The synthetic set is based on DIV2K-Val~\cite{zhang2018unreasonable}, where 3,000 images are randomly cropped to $512\times512$, and LR inputs are generated using the same Real-ESRGAN degradation process.
We further evaluate real-world performance on the RealSR dataset~\cite{cai2019toward}, which contains 100 paired LR–HR images at resolutions of $128\times128$ and $512\times512$.

\begin{table}[ht]
\centering
\vspace{-0.3cm}
\caption{Quantitative comparison of state-of-the-art Real-ISR methods. To evaluate the effectiveness of our alignment strategy, the best results among the base model (C-FLUX), its DP$^2$O-fine-tuned variant (DP$^2$O-FLUX), and our proposed method are highlighted with a \colorbox{lightred}{light red} background. The best and second best results across all methods are highlighted in \textcolor{red}{\textbf{red}} and \textcolor{blue}{\textbf{blue}}, respectively.}
\label{tab:sr_performance}
\resizebox{\textwidth}{!}{
\begin{tabular}{c | l | c c c c c c c c c}
\toprule
Datasets & Methods & PSNR$\uparrow$ & SSIM$\uparrow$ & LPIPS$\downarrow$ & MUSIQ$\uparrow$ & CLIPIQA$\uparrow$ & NIMA$\uparrow$ & TOPIQ-IAA$\uparrow$ & Q-Align$\uparrow$ & VQ-R1$\uparrow$ \\
\midrule
 & StableSR & 24.54 & 0.7004 & 0.3068 & 65.92 & 0.6325 & 4.7880 & 4.5738 & 3.2930 & 3.7670 \\
 & DiffBIR & 24.83 & 0.6501 & 0.3650 & 69.28 & 0.7053 & 4.9192 & 4.8973 & 3.7889 & 4.1030 \\
 & SeeSR & 25.14 & 0.7211 & 0.3007 & 69.82 & 0.6705 & 4.9196 & 4.8548 & 3.7190 & 3.9830 \\
 & OSEDiff & \textcolor{blue}{\textbf{25.15}} & \textcolor{blue}{\textbf{0.7341}} & 0.2921 & 69.09 & 0.6693 & 4.8953 & 4.7545 & 3.6962 & 4.0930 \\
 & PiSA-SR & \textcolor{red}{\textbf{25.50}} & \textcolor{red}{\textbf{0.7418}} & \textcolor{red}{\textbf{0.2672}} & 70.15 & 0.6698 & 4.8954 & 4.7431 & 3.6354 & 4.0180 \\
RealSR & TSD-SR & 23.40 & 0.6938 & \textcolor{blue}{\textbf{0.2823}} & 71.26 & \textcolor{red}{\textbf{0.7416}} & 5.0439 & 4.8958 & 3.8444 & 4.0800 \\
 & DIT4SR & 23.52 & 0.6683 & 0.3153 & 67.70 & 0.6305 & 4.8737 & 4.5709 & 3.3875 & 3.9430 \\
 & LucidFlux & 20.62 & 0.6016 & 0.3895 & \textcolor{blue}{\textbf{71.95}} & \textcolor{blue}{\textbf{0.7397}} & \textcolor{red}{\textbf{5.2414}} & \textcolor{red}{\textbf{5.1494}} & 3.7662 & 3.9010 \\
 & HYPIR & 22.76 & 0.6669 & 0.3180 & 65.89 & 0.6369 & 4.9407 & 4.7330 & 3.6814 & 4.1530 \\
\cmidrule{2-11}
 & C-FLUX & 24.42 & 0.6742 & 0.3376 & 68.70 & 0.6221 & 4.8799 & 4.6303 & 3.4943 & 3.9650 \\
 & DP$^2$O-FLUX & 24.51 & 0.6774 & 0.3259 & \cellcolor{lightred}\textcolor{red}{\textbf{72.42}} & \cellcolor{lightred}0.7156 & 4.9950 & 4.8873 & \textcolor{blue}{\textbf{3.9757}} & \textcolor{blue}{\textbf{4.2170}} \\
 & \textbf{Ours} & \cellcolor{lightred}24.68 & \cellcolor{lightred}0.7101 & \cellcolor{lightred}0.3202 & 71.70 & 0.7007 & \cellcolor{lightred}\textcolor{blue}{\textbf{5.0567}} & \cellcolor{lightred}\textcolor{blue}{\textbf{4.9533}} & \cellcolor{lightred}\textcolor{red}{\textbf{4.1596}} & \cellcolor{lightred}\textcolor{red}{\textbf{4.2930}} \\
\midrule
 & StableSR & 23.26 & 0.5728 & 0.3112 & 65.78 & 0.6767 & 4.9243 & 4.7608 & 3.5213 & 3.8925 \\
 & DiffBIR & 23.14 & 0.5441 & 0.3669 & 69.87 & 0.7298 & 5.2070 & \textcolor{blue}{\textbf{5.1431}} & 4.1015 & 4.2014 \\
 & SeeSR & 23.68 & 0.6043 & 0.3194 & 68.68 & 0.6936 & 5.0749 & 4.9928 & 3.9770 & 4.2013 \\
 & OSEDiff & \textcolor{blue}{\textbf{23.72}} & \textcolor{red}{\textbf{0.6109}} & 0.2942 & 67.97 & 0.6682 & 4.9664 & 4.8432 & 3.8358 & 4.1832 \\
 & PiSA-SR & \textcolor{red}{\textbf{23.87}} & \textcolor{blue}{\textbf{0.6058}} & \textcolor{blue}{\textbf{0.2823}} & 69.68 & 0.6929 & 4.9584 & 4.8572 & 3.8804 & 4.2054 \\
DIV2K-Val & TSD-SR & 22.41 & 0.5644 & \textcolor{red}{\textbf{0.2710}} & 71.64 & \textcolor{blue}{\textbf{0.7559}} & 5.1239 & 5.0116 & 4.0147 & 4.1529 \\
 & DIT4SR & 21.79 & 0.5487 & 0.3451 & 67.96 & 0.6619 & 5.0888 & 4.8515 & 3.7135 & 4.1551 \\
 & LucidFlux & 20.42 & 0.5106 & 0.3769 & 71.52 & \textcolor{red}{\textbf{0.7827}} & \textcolor{red}{\textbf{5.3984}} & \textcolor{red}{\textbf{5.3549}} & 4.0886 & 4.0852 \\
 & HYPIR & 22.17 & 0.5669 & 0.3073 & 64.73 & 0.6421 & 5.0178 & 4.8767 & 3.7544 & 4.0757 \\
\cmidrule{2-11}
 & C-FLUX & 22.71 & 0.5520 & 0.3349 & 69.70 & 0.6781 & 5.1084 & 4.9519 & 4.0573 & 4.2558 \\
 & DP$^2$O-FLUX & 22.72 & 0.5536 & \cellcolor{lightred}0.3216 & \cellcolor{lightred}\textcolor{red}{\textbf{72.98}} & \cellcolor{lightred}0.7536 & \cellcolor{lightred}\textcolor{blue}{\textbf{5.2482}} & \cellcolor{lightred}5.1347 & \textcolor{blue}{\textbf{4.3983}} & \textcolor{blue}{\textbf{4.4343}} \\
 & \textbf{Ours} & \cellcolor{lightred}22.88 & \cellcolor{lightred}0.5786 & 0.3260 & \textcolor{blue}{\textbf{71.95}} & 0.7302 & 5.1507 & 5.0695 & \cellcolor{lightred}\textcolor{red}{\textbf{4.4605}} & \cellcolor{lightred}\textcolor{red}{\textbf{4.4396}} \\
\bottomrule
\end{tabular}
}
\end{table}

\noindent\textbf{Experimental Setup.}
We compare our proposed method against a comprehensive suite of state-of-the-art generative SR approaches, including StableSR~\cite{wang2024exploiting}, DiffBIR~\cite{lin2024diffbir}, SeeSR~\cite{wu2024seesr}, OSEDiff~\cite{wu2024one}, PiSA-SR~\cite{sun2025pixel}, TSD-SR~\cite{dong2025tsd}, DIT4SR~\cite{duan2025dit4sr}, LucidFlux~\cite{fei2025lucidflux}, and HYPIR~\cite{lin2025harnessing}. 
To explicitly demonstrate the performance gains achieved by our alignment strategy, we further include the base model C-FLUX and its DPO-fine-tuned~\cite{wallace2024diffusion} variant DP$^2$O-FLUX (derived from DP$^2$O-SR~\cite{wu2025dp}) for comparison, evaluating all methods using their official default settings to ensure fairness. 
The performance is quantitatively assessed using a comprehensive set of FR metrics, including PSNR, SSIM~\cite{wang2004image}, and LPIPS~\cite{zhang2018unreasonable}, as well as NR metrics such as MUSIQ~\cite{ke2021musiq}, CLIPIQA~\cite{wang2023exploring}, Q-Align~\cite{wu2023q}, NIMA~\cite{talebi2018nima}, VQ-R1~\cite{wu2025visualquality}, and TOPIQ-IAA~\cite{chen2024topiq}.

\begin{wrapfigure}{r}{0.48\textwidth}
    \vspace{-10pt}
    \centering
    \includegraphics[width=\linewidth]{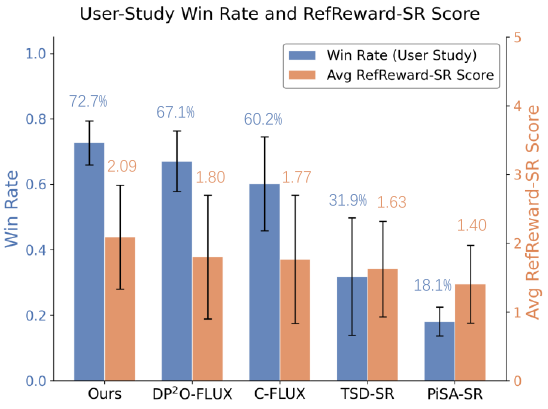}
    \caption{The result of the user study win rate and average RefReward-SR score.}
    \label{fig:user_study}
    \vspace{-10pt}
\end{wrapfigure}

\noindent\textbf{User Study.}
As demonstrated in Sec.~\ref{sec:refreward-result}, existing FR and NR metrics fail to fully capture human visual preferences in generative SR. Moreover, directly evaluating SR models using RefReward-SR may risk reward hacking~\cite{liu2025flow,xue2025dancegrpo}, since our method is optimized using this evaluator during reinforcement learning. To ensure an unbiased evaluation, we compare our method with the base model C-FLUX and its DPO-fine-tuned variant (DP$^2$O-FLUX) through a human preference study. Additionally, PiSA-SR and TSD-SR are included to examine RefReward-SR's generalization and alignment with human judgment on unseen methods.
Specifically, we randomly sample 10 images from DIV2K-Val and 10 images from RealSR to ensure both synthetic and real-world scenarios are covered. A total of 15 volunteers with prior experience in image processing are invited to participate in the evaluation. During each trial, participants are presented with one LR reference image and two HR reconstructions generated by different models. Annotators are asked to select the better reconstruction according to the same criteria adopted in RefSR-18K. Ties are allowed and counted as 0.5 wins when computing overall win rates.

\begin{figure}[t]
\centering
\includegraphics[width=0.9\textwidth]{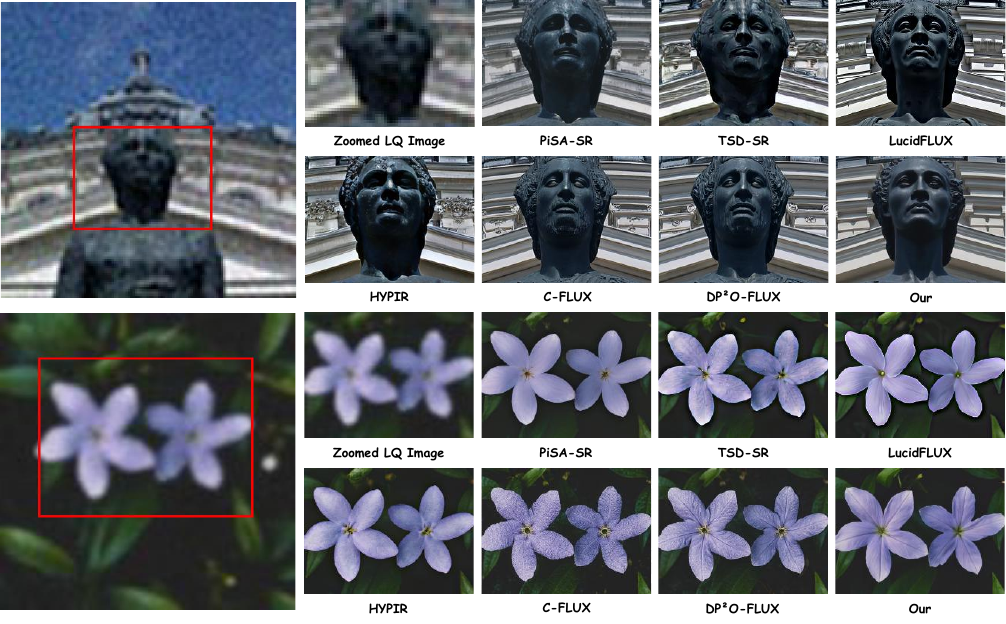}
\caption{Qualitative comparison of state-of-the-art generative SR methods. Please zoom in for a better view. Additional results are provided in the  \textit{Suppl. Mater.}.}
\label{fig:qual_comp}
\vspace{-10pt}  
\end{figure}

\noindent\textbf{Results and Analysis.}
As shown in Table~\ref{tab:sr_performance}, our method consistently improves upon the base model (C-FLUX). Compared with DP$^2$O-FLUX, which is also fine-tuned using a reinforcement learning paradigm (DPO), our approach maintains a leading position on the majority of metrics. This demonstrates that our training strategy effectively enhances low-level fidelity while simultaneously boosting low-level visual naturalness. Qualitative comparisons in Fig.~\ref{fig:qual_comp} further illustrate the advantages of our method. Compared with both the C-FLUX and the reinforcement learning baseline DP$^2$O-FLUX, our method produces visually cleaner, more realistic, and more semantically consistent high-resolution images, while remaining highly competitive with other recent SR methods. 

As illustrated in Fig.~\ref{fig:user_study}, the user study shows that our method achieves the highest win rate of 72.7\%, significantly outperforming DP$^2$O-FLUX (67.1\%) and C-FLUX (60.2\%). Meanwhile, the average RefReward-SR scores exhibit a strong positive correlation with the win rates, further validating the effectiveness of RefReward-SR as a reliable evaluation metric. Moreover, optimizing SR models with our proposed framework consistently drives the generation process toward better alignment with human semantic preferences. Interestingly, although C-FLUX performs worse than TSD-SR and PiSA-SR under conventional NR/FR metrics, it achieves a higher human preference win rate. This implies that existing metrics are insufficient for measuring alignment with human semantic preferences, which is consistent with the findings reported in Sec.~\ref{sec:refreward-result}.

\subsection{Ablation Study}
\label{sec:ablation}

\begin{figure}[t]
\centering
\begin{minipage}{0.48\textwidth}
\centering
\scriptsize
\captionof{table}{Ablation study of the RefReward-SR evaluator on both in-domain and out-of-domain test sets. Agreement is reported in \%.}
\label{tab:ablation_reward}
\setlength{\tabcolsep}{4pt}
\begin{tabular}{lcc}
\toprule
Variant & In-Domain & Out-of-Domain \\
\midrule
1K Data & $81.2 \pm 1.9$ & $76.6 \pm 3.2$ \\
2K Data & $82.7 \pm 1.9$ & $77.4 \pm 1.8$ \\
w/o G-L & $83.6 \pm 2.6$ & $78.1 \pm 2.3$ \\
\midrule
\textbf{Ours (Full)} & $\mathbf{85.0 \pm 2.4}$ & $\mathbf{80.2 \pm 2.2}$ \\
\bottomrule
\end{tabular}
\end{minipage}
\hfill
\begin{minipage}{0.48\textwidth}
\centering
\includegraphics[width=0.8\textwidth]{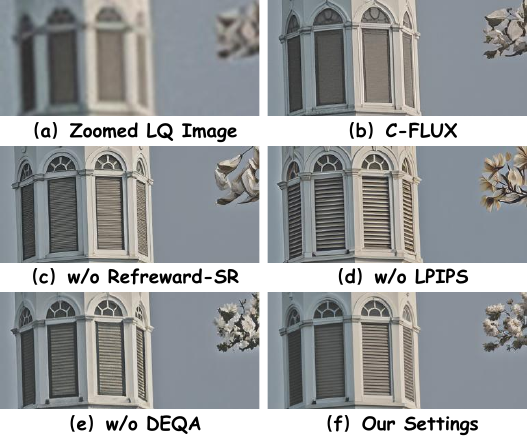}
\caption{Visual comparison of different reward components in the ablation study.}
\label{fig:ablation_visual}
\end{minipage}
\end{figure}

\noindent\textbf{Ablation on the RefReward-SR Evaluator.} As reported in Table~\ref{tab:ablation_reward}, we assess our reward model's key design choices using human alignment metrics across in-domain and out-of-domain test sets. We compare the full model against variants trained on reduced data ($1\text{K}$ and $2\text{K}$ groups) and without the global-local crop scoring mechanism. Scaling the training data yields consistent gains, demonstrating that larger preference datasets enhance human alignment and model generalization. Furthermore, removing crop scoring causes a performance drop, especially on the out-of-domain set, highlighting its effectiveness in capturing fine-grained semantic inconsistencies. Ultimately, the full configuration of \textbf{RefReward-SR} achieves the highest agreement, validating both data scaling and our global-local scoring design.

\noindent\textbf{Ablation on Preference-Aligned SR Optimization.} We investigate the contribution of each reward component defined in Eq.~\ref{eq:total_reward_grpo} during the GRPO fine-tuning of the C-FLUX model. As shown in Table~\ref{tab:ablation_sr} and Fig.~\ref{fig:ablation_visual}, we iteratively remove each reward term and evaluate the resulting reconstruction quality on the RealSR benchmark. Analysis of the experimental results reveals that each reward component plays a distinct role. When \textbf{RefReward-SR} is removed, the FR metrics experience only a slight decline. This is because \textbf{RefReward-SR} primarily focuses on high-level semantics, whereas content fidelity is mainly maintained by the LPIPS reward. Although some NR metrics even show improvement, this absence still leads to semantic distortion and structural twisting in the generated results. Removing the LPIPS reward causes a significant drop in low-level fidelity metrics. While the model can still maintain a rough semantic correspondence, the generated details exhibit increased deviation from the LR input. Without the DEQA reward, the related NR metrics drop significantly compared to the full model. However, due to the enhanced semantic plausibility, these metrics still show a relative increase over the base model. Visually, the clarity of the results decreases, and the details become blurry. Overall, the full model achieves the most balanced performance across all metrics, yielding the most satisfactory generated results.

\begin{table*}[t]
\centering
\caption{Ablation study of reward components on the RealSR benchmark. The best and second best results are highlighted in \textcolor{red}{\textbf{red}} and \textcolor{blue}{\textbf{blue}}, respectively.}
\label{tab:ablation_sr}
\resizebox{\textwidth}{!}{
\begin{tabular}{l | c c c c c c c c c}
\toprule
Methods & PSNR$\uparrow$ & SSIM$\uparrow$ & LPIPS$\downarrow$ & MUSIQ$\uparrow$ & CLIPIQA$\uparrow$ & NIMA$\uparrow$ & TOPIQ-IAA$\uparrow$ & Q-Align$\uparrow$ & VQ-R1$\uparrow$ \\
\midrule
C-FLUX (Base) & 24.42 & 0.6742 & 0.3376 & 68.70 & 0.6221 & 4.8799 & 4.6303 & 3.4943 & 3.9650 \\
\midrule
w/o RefReward-SR & \textcolor{blue}{\textbf{23.89}} & 0.6920 & 0.3211 & \textcolor{blue}{\textbf{73.48}} & \textcolor{blue}{\textbf{0.7239}} & \textcolor{blue}{\textbf{5.2534}} & \textcolor{blue}{\textbf{5.0935}} & \textcolor{blue}{\textbf{4.1656}} & 4.1655 \\
w/o LPIPS & 21.98 & 0.5798 & 0.4026 & \textcolor{red}{\textbf{74.55}} & \textcolor{red}{\textbf{0.7434}} & \textcolor{red}{\textbf{5.4195}} & \textcolor{red}{\textbf{5.2948}} & \textcolor{red}{\textbf{4.3521}} & \textcolor{red}{\textbf{4.3348}} \\
w/o DEQA & 23.87 & \textcolor{blue}{\textbf{0.7099}} & \textcolor{red}{\textbf{0.2967}} & 69.02 & 0.6153 & 5.0189 & 4.6552 & 3.5313 & 4.0188 \\
\midrule
\textbf{Ours (Full)} & \textcolor{red}{\textbf{24.68}} & \textcolor{red}{\textbf{0.7101}} & \textcolor{blue}{\textbf{0.3202}} & 71.70 & 0.7007 & 5.0567 & 4.9533 & 4.1596 & \textcolor{blue}{\textbf{4.2930}} \\
\bottomrule
\end{tabular}
}
\vspace{-0.4cm}
\end{table*}

\section{Conclusion}
\label{sec:conclusion}

In this paper, we investigate the limitations of existing super-resolution (SR) evaluation metrics in capturing semantic understanding, which often leads to misalignment with human perceptual preferences. To address this issue, we introduce RefSR-18K, the first large-scale LR-conditioned preference dataset for generative SR with explicit annotations on semantic consistency and plausibility. Based on this dataset, we propose RefReward-SR, an MLLM-based evaluator designed to assess SR results from a semantic-level human preference perspective. We further incorporate RefReward-SR into the SR optimization pipeline as a core reward signal, enabling preference-aligned SR generation beyond conventional ground-truth distribution matching. Extensive experiments demonstrate that RefReward-SR provides a reliable proxy for human evaluation, and that the resulting preference-aligned SR model effectively suppresses generative artifacts while improving semantic fidelity and visual naturalness. We hope this work will inspire future research toward more human-centric evaluation and optimization paradigms for low-level vision tasks.


%
%
\bibliographystyle{splncs04}
\bibliography{main}

\clearpage
\appendix

\section{Details on RefSR-18K: Construction and Analysis}
\label{sec:refsr-18k}

\subsection{RefSR-18K Dataset Annotation Guidelines}

\begin{figure}[htbp]
  \centering
  \includegraphics[width=\textwidth]{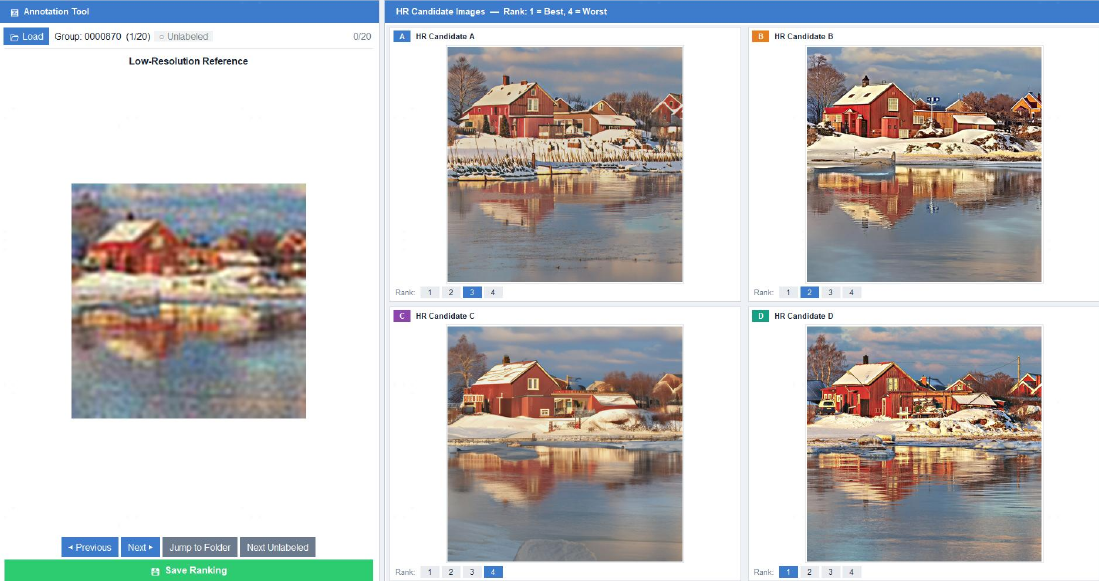}
  \caption{User interface of our annotation platform. The left panel shows the LR reference as a semantic anchor, while the right displays four HR candidates for annotators to rank from 1 (best) to 4 (worst).}
  \label{fig:annotation_tool}
\end{figure}

To accurately capture human perceptual preferences for generative super-resolution (SR) results, we have designed a rigorous relative quality ranking protocol. In each evaluation trial, the system will present annotators with a low-resolution (LR) reference image alongside four high-resolution (HR) candidate images reconstructed by different generative SR algorithms. Fig.~\ref{fig:annotation_tool} illustrates the user interface of our dedicated annotation platform.

Treating the LR image as a semantic anchor, the annotator's core task is to evaluate the HR candidates across two dimensions: \textbf{LR-HR semantic consistency} and \textbf{semantic plausibility}. Given that these two dimensions are highly correlated in actual human visual perception (e.g., severe structural distortions often simultaneously lead to semantic shifts), this guideline does not require separate scoring. Instead, annotators must output a global preference ranking from 1 to 4 (1 indicating the best perceptual quality, 4 indicating the worst) based on a holistic trade-off. The specific evaluation guidelines are as follows:


\begin{quote}
\ttfamily
\raggedright

\noindent \textbf{Dimension 1: LR-HR Semantic Consistency}

This dimension focuses on evaluating whether the HR reconstruction is faithful to the original semantic content of the LR input, requiring the model not to exhibit excessive "hallucinations" or arbitrarily alter image details. Annotators should focus on whether the HR image exhibits the following behaviors that severely violate consistency:

\begin{itemize}[label=\textbullet, leftmargin=1.5em]
\item \textbf{Object Addition/Deletion and Structural Alteration}: Compared to the LR reference image, whether new objects are fabricated out of thin air in the HR image, or original key structures are erased.
\item \textbf{Semantic Identity Shift}: Whether the core semantics of the target object have fundamentally changed. For example, erroneously reconstructing a blurry "tiger" contour in the LR image as a "cat", or "hallucinating" a blurry real text area into meaningless random patterns.
\item \textbf{Physical and Logical Inconsistencies}: Whether materials or local logic are incorrectly replaced. For example, reconstructing a brick wall in the LR as a smooth concrete wall, or generating a wooden fence as metal railings.
\item \textbf{Evaluation Criterion}: The more faithful the detail reconstruction is to the structure and semantic cues of the LR reference image, without arbitrarily "hallucinating" incorrect content, the better the candidate image performs in this dimension.
\end{itemize}

\vspace{1em}
\noindent \textbf{Dimension 2: Semantic Plausibility}

This dimension aims to evaluate whether the generated HR image conforms to real-world visual common sense, both globally and locally. Breaking away from rigid pixel-to-pixel comparison, focus on identifying inherent artifacts and distortions of generative models:

\begin{itemize}[label=\textbullet, leftmargin=1.5em]
\item \textbf{Local Distortions and Generative Artifacts}: Carefully check if the image contains obvious unnatural phenomena such as twisted window shapes, distorted human facial features, collapsed object edge structures, or irregular line jitters.
\item \textbf{Global Logical Violations}: Whether the scene content violates physical common sense, or if poor generation quality results in a completely unrecognizable block of colors, lacking the texture that a real natural image should have.
\item \textbf{Evaluation Criterion}: The closer the visual appearance of the image is to a real high-definition photograph, and the more naturally smooth and distortion-free the scene is, the better the candidate image performs in this dimension.
\end{itemize}

\vspace{1em}
\noindent \textbf{Holistic Preference Ranking Mechanism}

After independently considering the above two dimensions, annotators need to perform a final global quality aggregation to provide a ranking from 1 to 4.

\begin{itemize}[label=\textbullet, leftmargin=1.5em]
\item \textbf{Holistic Trade-off Decision}: The final ranking is not a simple linear weighting of the two dimensions, but a comprehensive judgment based on visual perception. Evaluations often face a trade-off between semantic plausibility and semantic consistency: for example, choosing between an image with realistic details but deviated semantics (high plausibility, low consistency) and one faithful to the original but containing artifacts (high consistency, low plausibility). Annotators need to judge which distortion is less acceptable based on true visual preferences to select the globally optimal reconstruction result.
\item \textbf{Tie-Handling Protocol}: To ensure sufficient discriminative power in the preference data, the annotation system defaults to requiring a strict total ordering. Tied rankings (e.g., 1, 1, 3, 4 or 2, 2, 3, 4) are permitted if and only if the reconstruction quality and visual appearance of multiple HR candidates are extremely close, making it impossible for the human visual system to reliably distinguish their superiority or inferiority.
\end{itemize}

\end{quote}

\subsection{Dataset Analysis}
In this subsection, we systematically analyze the constructed RefSR-18K dataset from two key perspectives: its semantic diversity across visual scenarios and the statistical distribution of the collected human preference rankings.

\begin{wrapfigure}{r}{0.48\textwidth}
    \vspace{-20pt}
    \centering
    \includegraphics[width=\linewidth]{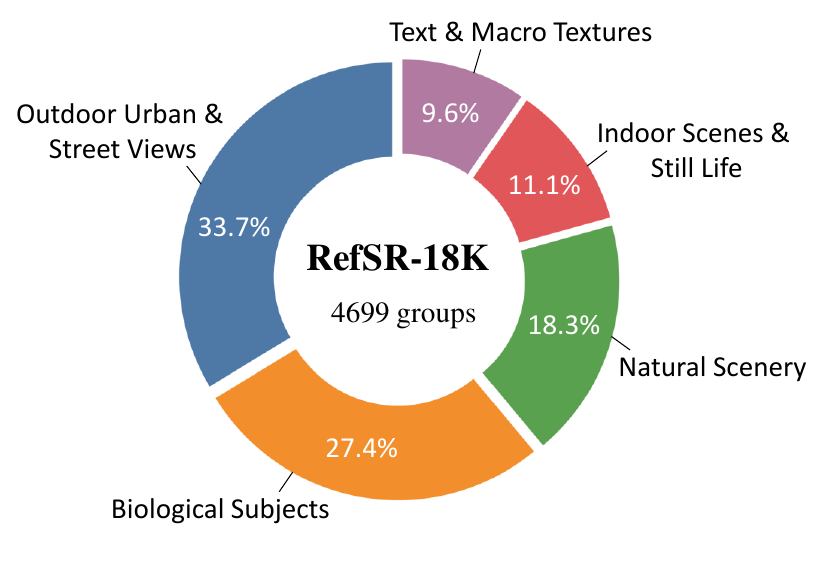}
    \caption{Semantic categorization of the RefSR-18K dataset into five major groups.}
    \label{fig:pie_chart}
    \vspace{-20pt}
\end{wrapfigure}

\noindent\textbf{Semantic Categorization.}
To ensure the robustness and generalization of the proposed reward model, RefSR-18K is curated to encompass a broad spectrum of visual scenarios. We leverage Qwen3-VL-8B~\cite{Qwen3-VL} to systematically analyze and categorize the 4,699 high-quality annotated samples into five major groups based on their dominant semantic content. As illustrated in Fig.~\ref{fig:pie_chart}, our dataset reflects the inherent complexity of real-world visual environments—ranging from expansive outdoor landscapes and structured architectural scenes to intricate indoor details, essential biological features, and challenging local textures. By maintaining such extensive visual coverage through this automated yet precise categorization, we effectively prevent the reward model from developing scene-specific biases and ensure its reliable performance across diverse SR applications.

\begin{wrapfigure}{r}{0.48\textwidth}
    \vspace{-15pt}
    \centering
    \includegraphics[width=\linewidth]{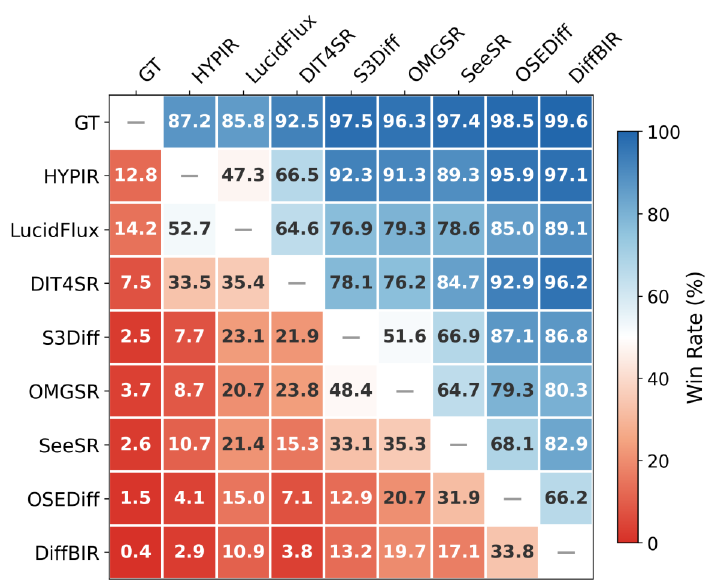}
    \caption{Pairwise win-rate matrix among evaluated HR candidates.}
    \label{fig:win_rate_matrix}
    \vspace{-30pt}
\end{wrapfigure}

\noindent\textbf{Preference Statistics and Win Rates.}
Based on the collected holistic preference rankings (from 1 to 4), we compute the pairwise win rates among evaluated HR candidates, as illustrated in Figure~\ref{fig:win_rate_matrix}. A detailed analysis of the win-rate matrix reveals two notable phenomena regarding human visual perception.

First, the Ground Truth (GT) is not universally preferred. When compared with the most advanced generative SR models, human evaluators favor the reconstructed images over the actual GT in over 10\% of the comparisons (e.g., 12.8\% against HYPIR and 14.2\% against LucidFlux). This suggests that state-of-the-art generative models can occasionally synthesize results that are perceptually preferred by humans at the semantic level over the original camera-captured images, a similar phenomenon observed by Chen et al. \cite{chen2025toward}.

Second, although HYPIR exhibits relatively lower No-Reference (NR) and Full-Reference (FR) scores in Table 2, it achieves competitive win rates against other baselines, with its overall perceptual performance ranking second only to GT. This subjective preference can be largely attributed to its capability in preserving semantically realistic and natural details. This phenomenon further suggests that relying exclusively on traditional NR and FR metrics may have certain limitations when assessing the true visual quality of generative SR methods.

\section{More Details on RefReward-SR}
\label{sec:supp_refreward}

\subsection{Detailed Prompt Design}
To effectively align the MLLM evaluator with human visual preferences, we meticulously designed a reference-aware instruction prompt. The prompt casts the MLLM as a highly discerning expert and mandates a rigorous step-by-step visual reasoning process. The exact prompt $c$ used in RefReward-SR is as follows:

\begin{quote}
\ttfamily
\raggedright
You are a highly discerning expert in image quality assessment.
Your task is to evaluate a super-resolution (SR) output by comparing it with the original low-resolution reference.
You will be given two images: the first (LR) shows the original scene, and the second (HR) is the super-resolution result.
Assume the HR image may contain problematic regions.
Carefully inspect both images and identify any areas in the HR image that look unnatural or distorted (e.g., distorted regions or warped lines, unrealistic textures, unreasonable objects) or semantically inconsistent with the LR image (e.g., missing, added, or altered objects, inconsistent texture).
For each problematic region, state its approximate location in the HR image (e.g., 'top left', 'center lower') and briefly explain why it appears incorrect.
After inspection, assign an overall quality score from 1.00 to 5.00, with two decimal places (e.g., 1.31, 2.77, 4.53).
First output your reasoning in <thinking>...</thinking> tags (a brief summary of observed defects and semantic issues).
Then output only one numeric score in <answer>...</answer> tags (no extra text).
\end{quote}

\subsection{Evaluation Metrics}
\label{sec:supp_metrics}

To quantitatively measure the alignment between the predicted quality rankings produced by RefReward-SR and the human-annotated ground-truth rankings, we adopt three complementary metrics: \textbf{Agreement}, \textbf{Recall@1}, and \textbf{Filter@1}.

\noindent\textbf{Score-to-Rank Conversion.}
RefReward-SR outputs a continuous quality score $s_i$ for each HR candidate $i$. To compare with the discrete human rankings, we convert these scores into ordinal ranks via descending sort. Ties are resolved using mid-ranks: if multiple candidates share the same score and jointly occupy rank positions $[a, b]$, each is assigned the average rank $(a+b)/2$. For example, two candidates tied for 2nd and 3rd place both receive a rank of $2.5$.

\noindent\textbf{Agreement.}
An agreement measures the consistency of pairwise preference orderings between the predicted and ground-truth rankings. Given $N=4$ HR candidates per image group, we enumerate all $\binom{4}{2}=6$ pairwise comparisons. For each pair $(i, j)$, the order relation is encoded as:
\begin{equation}
    o(i,j) = \begin{cases}
        +1, & \text{if } r_i > r_j \text{ (candidate } i \text{ is ranked worse),} \\
        -1, & \text{if } r_i < r_j \text{ (candidate } i \text{ is ranked better),} \\
        \phantom{+}0, & \text{if } r_i = r_j \text{ (tied),}
    \end{cases}
\end{equation}
where $r_i$ denotes the rank of candidate $i$. Let $o^{\text{pred}}$ and $o^{\text{gt}}$ denote the pairwise order functions derived from the predicted and ground-truth rankings, respectively. The Agreement rate is computed as:
\begin{equation}
    \text{Agreement} = \frac{1}{|\mathcal{P}|} \sum_{(i,j) \in \mathcal{P}} \mathbb{1}\big[o^{\text{pred}}(i,j) = o^{\text{gt}}(i,j)\big],
\end{equation}
where $\mathcal{P}$ is the set of all ordered pairs across all evaluated image groups. This metric primarily reflects the correctness of the entire preference ordering, including the handling of ties.

\noindent\textbf{Recall@1.}
Recall@1 measures the proportion of image groups in which the human-annotated best candidate (\emph{i.e.}, the one with rank $1$) is also ranked first by the model. Formally:
\begin{equation}
    \text{Recall@1} = \frac{1}{|\mathcal{G}|} \sum_{g \in \mathcal{G}} \mathbb{1}\big[\mathcal{B}^{\text{gt}}_g \cap \mathcal{B}^{\text{pred}}_g \neq \emptyset\big],
\end{equation}
where $\mathcal{G}$ denotes the set of all image groups, and $\mathcal{B}^{\text{gt}}_g$, $\mathcal{B}^{\text{pred}}_g$ are the sets of candidates assigned the best rank (\emph{i.e.}, the minimum rank value) in the ground truth and prediction, respectively. 

\noindent\textbf{Filter@1.}
Complementary to Recall@1, Filter@1 measures the proportion of image groups in which the human-annotated worst candidate (\emph{i.e.}, the one with the maximum rank value) is also ranked last by the model:
\begin{equation}
    \text{Filter@1} = \frac{1}{|\mathcal{G}|} \sum_{g \in \mathcal{G}} \mathbb{1}\big[\mathcal{W}^{\text{gt}}_g \cap \mathcal{W}^{\text{pred}}_g \neq \emptyset\big],
\end{equation}
where $\mathcal{W}^{\text{gt}}_g$ and $\mathcal{W}^{\text{pred}}_g$ are the sets of candidates assigned the worst rank in the ground truth and prediction, respectively. 

\subsection{Fast Inference via No-Think Mode}
\label{sec:supp_no_think}

The complete visual reasoning evaluation described above requires generating lengthy explanatory trajectories within the \texttt{<thinking>...</thinking>} tags, which introduces significant computational overhead. When RefReward-SR is deployed as a reward function in reinforcement learning, it must evaluate a massive number of candidate images at each training step, making inference efficiency a critical bottleneck.

To address this issue, we introduce a fast inference mode~\cite{guo2025imagedoctor} that bypasses the explicit reasoning process while preserving the model's scoring capability. Specifically, after constructing the standard input prompt, we append a pre-filled placeholder \texttt{<thinking>...</thinking><answer>} directly to the model's input sequence before generation. This causes the model to assume that the thinking phase has already been completed, prompting it to directly output the numeric quality score and thereby accelerating inference.

We compare the performance of this ``no-think'' mode against the default mode in Table~\ref{tab:no_think_results}. The results indicate that the fast inference mode achieves highly consistent performance across all metrics compared to the default reasoning mode.

\begin{table}[!ht]
\centering
\caption{Quantitative comparison (in \%) between the default reasoning mode and the fast inference (No-Think) mode of RefReward-SR. All $\pm$ values indicate the maximum deviation from the mean across different annotator rankings.}
\label{tab:no_think_results}
\resizebox{0.9\textwidth}{!}{
\begin{tabular}{lcccccc}
\toprule
\textbf{Method} & \multicolumn{3}{c}{\textbf{In-Domain}} & \multicolumn{3}{c}{\textbf{Out-of-Domain}} \\
\cmidrule(lr){2-4} \cmidrule(lr){5-7}
& \textbf{Agreement} & \textbf{Recall@1} & \textbf{Filter@1} & \textbf{Agreement} & \textbf{Recall@1} & \textbf{Filter@1} \\
\midrule
RefReward-SR (Default) & $\mathbf{85.0 \pm 2.4}$ & $\mathbf{84.5}$ & $78.0$ & $\mathbf{80.2 \pm 2.2}$ & $\mathbf{77.0}$ & $\mathbf{73.0}$ \\
RefReward-SR (No-Think) & $84.9 \pm 2.1$ & $\mathbf{84.5}$ & $\mathbf{79.0}$ & $79.8 \pm 2.3$ & $\mathbf{77.0}$ & $\mathbf{73.0}$ \\
\bottomrule
\end{tabular}
}
\end{table}

\subsection{Detailed Pipeline for Global-Local Crop Filtering}
\label{sec:supp_crop_filtering}

As introduced in the main text, we employ a semantic-guided pipeline to extract informative local regions. Here, we provide the specific implementation details of the bounding box detection and multi-stage filtering mechanisms.

\noindent\textbf{Semantic Region Detection.}
Given a ground-truth (GT) image, we employ the Recognize Anything Model (RAM)\cite{zhang2023recognize} to generate a comprehensive set of open-vocabulary semantic tags describing the image content. These tags are then fed as text prompts to Grounding DINO\cite{liu2023grounding}, which produces a set of candidate bounding boxes with associated confidence scores. This enables fully automatic, class-agnostic detection of semantically meaningful regions without requiring manual annotation.

\noindent\textbf{Multi-Stage Box Filtering.}
To balance inference efficiency and evaluation effectiveness, the raw candidate boxes are refined through a precise filtering pipeline to retain only regions suitable for fine-grained local quality evaluation:
\begin{enumerate}[leftmargin=1.5em]
    \item \textbf{Confidence \& NMS Filtering}: Candidate boxes with detection confidence below a threshold $\tau_{\text{box}}=0.25$ are discarded. Standard Non-Maximum Suppression (NMS) with an IoU threshold of $0.5$ is then applied to remove redundant overlapping detections.
    \item \textbf{Low-Texture Region Exclusion}: Boxes whose semantic labels match a predefined set of low-texture or homogeneous categories (\emph{e.g.}, sky, clouds, water surfaces, snow, fog) are removed, as these areas provide a limited discriminative signal for assessing generative SR quality.
    \item \textbf{Geometric Constraints}: Boxes with extreme aspect ratios ($\max(w/h,\, h/w)$ > 4.5) are discarded to avoid elongated, non-informative strips. Additionally, boxes whose normalized area falls outside the range $[0.1, 0.7]$ of the total image area are filtered out---eliminating both trivially small patches and near-full-image regions that would duplicate the global assessment.
    \item \textbf{Overlap Deduplication}: Among remaining boxes with mutual overlap exceeding $70\%$ ($\text{IoU} > 0.7$), only the larger box is retained to maximize spatial coverage while avoiding redundant evaluations.
    \item \textbf{Diversity-Aware Top-$K$ Selection}: When the number of surviving boxes exceeds a maximum budget of $K=4$, we apply a diversity-aware selection strategy. Specifically, we compute the mean pairwise IoU of each box with all others, and greedily select the $K$ boxes with the lowest mean IoU. This balances evaluation efficiency while maximizing the coverage of diverse semantic details across the image.
\end{enumerate}

Ultimately, the final selected local crops are evaluated by the RefReward-SR, and their scores are aggregated with the global score using the area-weighted formula detailed in the main text.

\section{Implementation Details}
\label{sec:supp_implementation}

All experiments are conducted on a single node equipped with 8$\times$ NVIDIA H20 GPUs. To ensure full reproducibility, we detail the training configurations for both stages below.

\subsection{Training Configuration for RefReward-SR}

We fine-tune RefReward-SR from the pre-trained Qwen3-VL-8B-Instruct~\cite{Qwen3-VL} checkpoint using Group Relative Policy Optimization (GRPO). The training is distributed across all 8 GPUs via \texttt{torchrun} with DeepSpeed ZeRO Stage~3 for memory-efficient optimization. Key hyperparameters are summarized in Table~\ref{tab:refreward_hparams}.

\begin{table}[!ht]
\centering
\caption{Training hyperparameters for RefReward-SR.}
\label{tab:refreward_hparams}
\begin{tabular}{ll}
\toprule
\textbf{Hyperparameter} & \textbf{Value} \\
\midrule
Base model & Qwen3-VL-8B-Instruct \\
Optimization strategy & DeepSpeed ZeRO Stage 3 \\
Number of GPUs & 8 $\times$ H20 \\
Precision & BFloat16 \\
Attention implementation & FlashAttention-2 \\
Max prompt length & 1,024 \\
Number of generations per group & 6 \\
Per-device image group size & 1 \\
Per-device image pair size & 4 \\
Gradient accumulation steps & 2 \\
Number of training steps & 1200 \\
Data seed & 42 \\
\bottomrule
\end{tabular}
\end{table}

\subsection{Training Configuration for GRPO-based SR Fine-tuning}

For the downstream SR model, we fine-tune the C-FLUX model from DP$^2$O-SR~\cite{wu2025dp} using LoRA-based GRPO. LoRA adapters are injected exclusively into the attention layers (\texttt{to\_q}, \texttt{to\_k}, \texttt{to\_v}, \texttt{to\_out}) of the FLUX DiT transformer, while both the ControlNet branch and the VAE remain entirely frozen throughout training. The reward signal is a composite of three complementary objectives: (1)~\textbf{RefReward-SR} (our trained reward model, operating in no-think mode), (2)~\textbf{LPIPS}~\cite{zhang2018unreasonable}, and (3)~\textbf{DeQA}~\cite{you2025teaching}, each weighted equally at $1.0$. The complete set of hyperparameters is listed in Table~\ref{tab:grpo_sr_hparams}.

\begin{table}[!ht]
\centering
\caption{Training hyperparameters for preference-aligned SR via GRPO.}
\label{tab:grpo_sr_hparams}
\begin{tabular}{ll}
\toprule
\textbf{Hyperparameter} & \textbf{Value} \\
\midrule
\multicolumn{2}{l}{\textit{Model \& Architecture}} \\
Base model & C-FLUX (DP$^2$O-SR~\cite{wu2025dp}) \\
ControlNet & Frozen \\
VAE & Frozen \\
LoRA target modules & QKV \& output projections of DiT Transformer \\
LoRA rank / alpha & 32 / 64 \\
ControlNet conditioning scale & 1.0 \\
\midrule
\multicolumn{2}{l}{\textit{Optimization}} \\
Number of GPUs & 8 $\times$ H20 \\
Precision & BFloat16 \\
Learning rate & $3 \times 10^{-4}$ \\
Weight decay & $1 \times 10^{-4}$ \\
Max gradient norm & 1.0 \\
Train batch size (per device) & 2 \\
Gradient accumulation steps & 12 \\
Max training steps & 180 \\
Seed & 42 \\
\midrule
\multicolumn{2}{l}{\textit{Sampling \& GRPO}} \\
Image resolution & $512 \times 512$ \\
Sampling steps & 25 \\
Guidance scale & 2.5 \\
Number of generations per group & 12 \\
PPO clip range ($\epsilon$) & $1 \times 10^{-4}$ \\
Advantage clip max & 5.0 \\
$\eta$ (KL penalty) & 0.3 \\
Sampler seed & 1,223,627 \\
\midrule
\multicolumn{2}{l}{\textit{Reward Weights}} \\
$\lambda_{\text{RefReward-SR}}$ & 1.0 \\
$\lambda_{\text{DeQA}}$ & 1.0 \\
$\lambda_{\text{LPIPS}}$ & 1.0 \\
\bottomrule
\end{tabular}
\end{table}

\section{More Visual Results}
\label{sec:supp_qualitative}

\begin{figure}[htbp]
  \centering
  \includegraphics[width=0.9\textwidth]{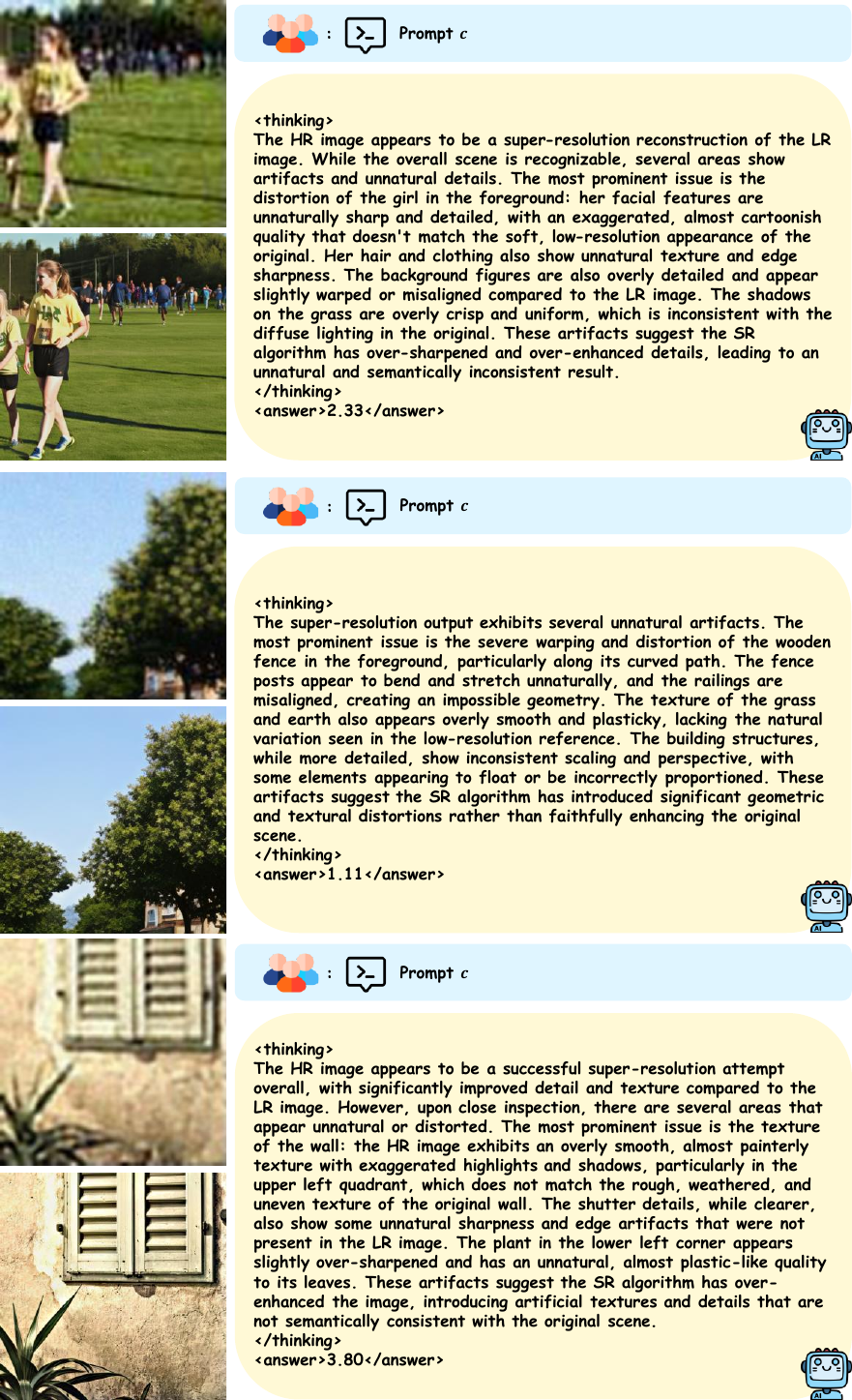}
  \caption{Additional visual results of RefReward-SR evaluations, illustrating the visual reasoning trajectories and the predicted quality scores.}
  \label{fig:supp_eval}
\end{figure}

\begin{figure}[htbp]
  \centering
  \includegraphics[width=\textwidth]{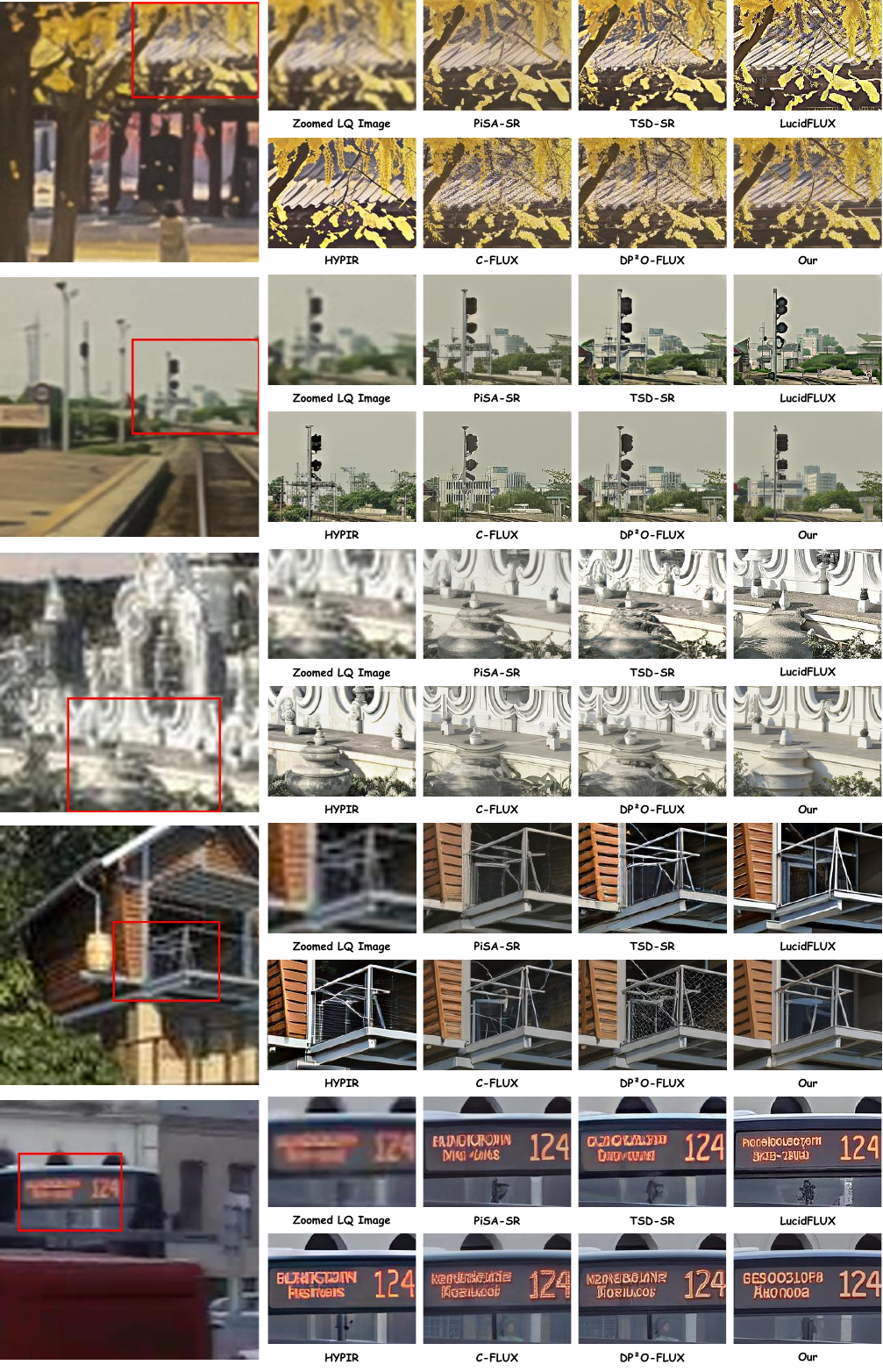}
  \caption{Visual comparisons of preference-aligned SR results on real-world datasets. Please zoom in for a better view.}
  \label{fig:supp_real}
\end{figure}

\begin{figure}[htbp]
  \centering
  \includegraphics[width=\textwidth]{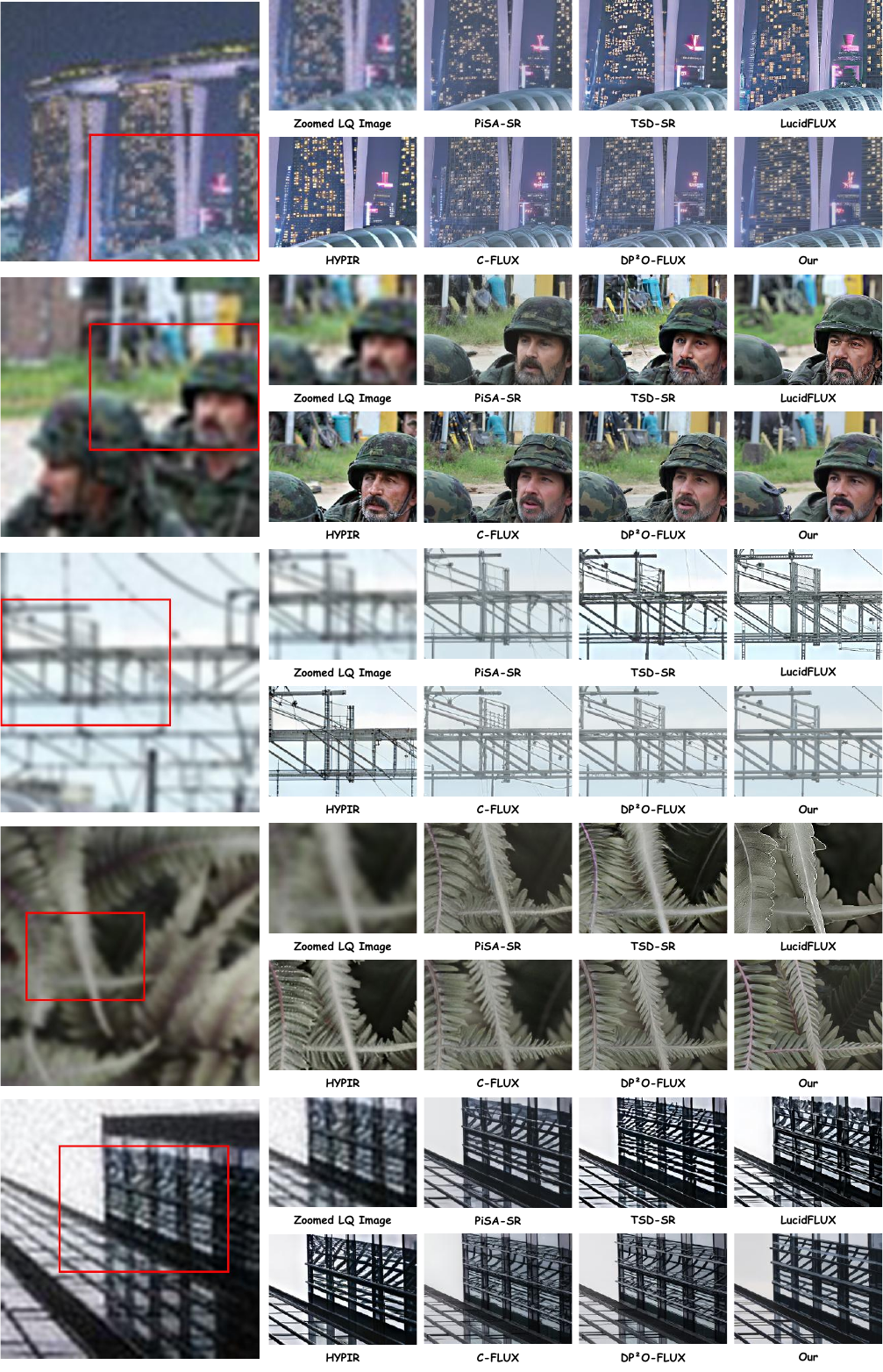}
  \caption{Visual comparisons of preference-aligned SR results on the DIV2K-val dataset. Please zoom in for a better view.}
  \label{fig:supp_div2k}
\end{figure}

We provide additional visual results for the RefReward-SR evaluations, alongside preference-aligned SR outputs on real-world images~\cite{cai2019toward,ai2024dreamclear} and DIV2K-val~\cite{zhang2018unreasonable} images in Figure~\ref{fig:supp_eval}, Figure~\ref{fig:supp_real}, and Figure~\ref{fig:supp_div2k}, respectively.

\section{Limitations}
\label{sec:limitations}

In this section, we discuss some limitations identified during the development of RefReward-SR.

\noindent\textbf{Dataset scale and model diversity.} Although our RefSR-18K dataset encompasses a wide variety of representative SR models and ground-truth images, and our experiments demonstrate RefReward-SR's excellent performance in both in-domain and out-of-domain scenarios, a continuously expanding dataset is still needed for more comprehensive RM training. Given the rapid evolution of generative AI, incorporating an even broader array of cutting-edge SR algorithms would allow RefReward-SR to generalize better and reach its full potential.

\noindent\textbf{Reward model training and reasoning.} During the training of RefReward-SR, we rely on reinforcement learning to guide the model to output its reasoning process. While our model can reasonably and accurately articulate the visual artifacts and semantic inconsistencies present in HR images, given the absence of explicit supervision on these reasoning trajectories during training, the full potential of its interpretable reasoning remains to be further unlocked. A promising direction is to augment the annotation process with human-written scoring rationales and perform an initial supervised fine-tuning (SFT) stage before reinforcement learning, which may lead to more reliable explanations and stronger alignment with human judgments.

\noindent\textbf{Granularity of reward for SR optimization.} In the current process of aligning SR models with human preferences, the reward signal is primarily utilized at the image level. While effective, this holistic approach might overlook the opportunity for more fine-grained penalties. Building upon our global-local crop scoring mechanism, it would be an interesting direction to explicitly identify the worst-performing local regions and apply targeted reinforcement learning optimization at the patch level.

\end{document}